\documentclass[12pt]{l4dc2023}

\usepackage{times}
\usepackage{amssymb}
\usepackage{amsmath}
\usepackage{hyperref}            
\usepackage{cleveref}
\usepackage{booktabs}  
\usepackage{caption}
\usepackage{graphicx}
\usepackage{multirow}
\usepackage[T1]{fontenc}
\usepackage{siunitx}

\usepackage{listings}
\newcommand{\Tau}{\makebox{\huge\ensuremath{\tau}}}

\usepackage{glossaries}
\glsdisablehyper
\newacronym{nn}{NN}{Neural Network}
\newacronym{rfi}{RFI}{Random Force Injection}
\newacronym{erfi}{ERFI}{Extended Random Force Injection}
\newacronym{rao}{RAO}{Random Actuation Offset}
\newacronym{rl}{RL}{Reinforcement Learning}
\newacronym{drl}{DRL}{Deep-Reinforcement Learning}
\newacronym{dm}{DM}{Domain Randomization}
\newacronym{td3}{TD3}{Twin Delayed Deep Deterministic}
\newacronym{ppo}{PPO}{Proximal Policy Optimization}
\newacronym{gae}{GAE}{Generalized Advantage Estimation}
\newacronym{sea}{SEA}{Series Elastic Actuator}
\newacronym{mlp}{MLP}{Multi-Layer Perceptron}
\newacronym{erfic}{ERFI-C}{Extended Random Force Injection Cumulative}
\newacronym{erfi50}{ERFI-50}{Extended Random Force Injection 50\%}
\newacronym{com}{CoM}{Center of Mass}
\newacronym{mdp}{MDP}{Markov Decision Process}
\newacronym{vae}{VAE}{Variational Autoencoder}
\newacronym{imu}{IMU}{Inertial Measurement Unit}

\title[Roll-Drop]{Roll-Drop: accounting for observation noise with a single parameter}
\author{%
 \Name{Luigi Campanaro} \Email{luigi@robots.ox.ac.uk}\\
 \Name{Daniele De Martini} \Email{daniele@robots.ox.ac.uk}\\
 \Name{Siddhant Gangapurwala} \Email{siddhant@robots.ox.ac.uk}\\
 \Name{Wolfgang Merkt} \Email{wolfgang@robots.ox.ac.uk}\\
 \Name{Ioannis Havoutis} \Email{ioannis@robots.ox.ac.uk}\\
 \addr Department of Engineering Science, University of Oxford
}

\begin{document}

\maketitle

\begin{abstract}%
This paper proposes a simple strategy for sim-to-real in \gls{drl} -- called Roll-Drop -- that uses dropout during simulation to account for observation noise during deployment without explicitly modelling its distribution for each state.
\Gls{drl} is a promising approach to control robots for highly dynamic and feedback-based manoeuvres, and accurate simulators are crucial to providing cheap and abundant data to learn the desired behaviour.
Nevertheless, the simulated data are noiseless and generally show a distributional shift that challenges the deployment on real machines where sensor readings are affected by noise.
The standard solution is modelling the latter and injecting it during training; while this requires a thorough system identification, Roll-Drop enhances the robustness to sensor noise by tuning only a single parameter.
We demonstrate an 80\% success rate when up to 25\% noise is injected in the observations, with twice higher robustness than the baselines.
We deploy the controller trained in simulation on a Unitree A1 platform and assess this improved robustness on the physical system.
Additional resources at: \url{https://sites.google.com/oxfordrobotics.institute/roll-drop}
\end{abstract}

\glsresetall


\begin{figure}[ht!]
    \centering
    \includegraphics[width=0.9\textwidth]{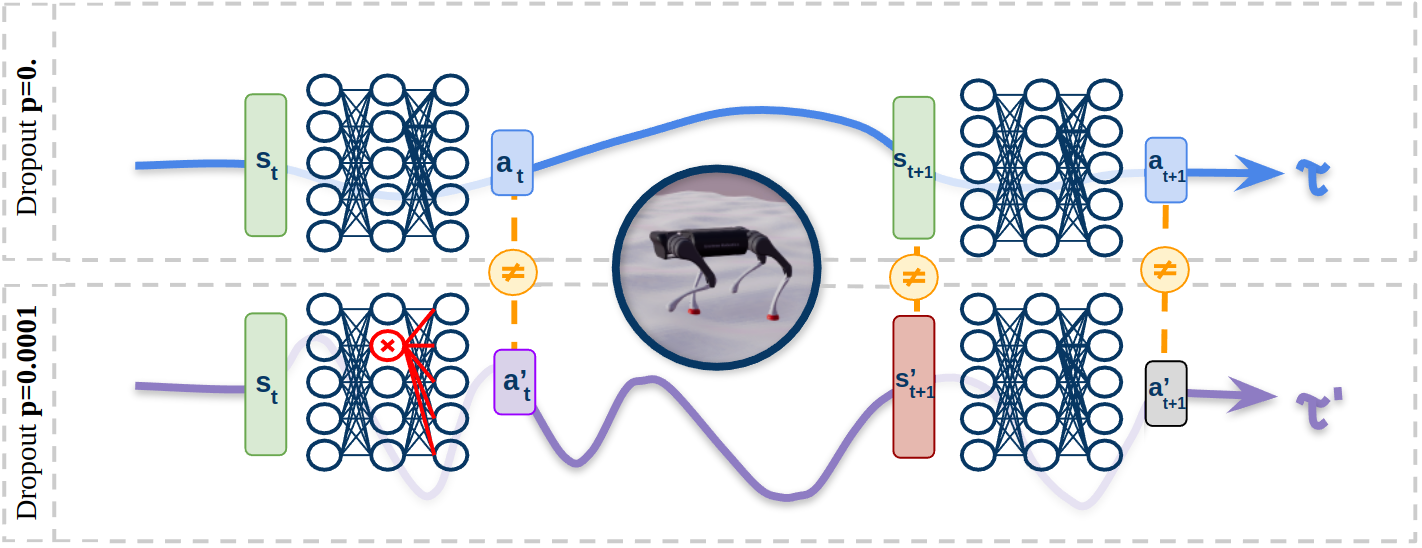}
    \caption{
    We present two policies, $\hat{\pi}$ trained with dropout during rollout (bottom) and $\pi$ without it (top).
    As the time-step $t$ precedes the occurrence of the first dropout and the training adopts the same random seed, both policies visited the same states and actions.
    After the first dropout is triggered, the policies will follow different trajectories, $\Tau_{\pi}$ and $\Tau'_{\hat{\pi}}$: [$a_t,..,a_T]_{\pi} \neq [a'_t,..,a'_T]_{\hat{\pi}}$, $[r_{t},..,r_{T}]_{\pi} \neq [r'_{t},..,r'_{T}]_{\hat{\pi}}$, and $[s_{t+1},..,s_{T}]_{\pi} \neq [s'_{t+1},..,s'_{T}]_{\hat{\pi}}$. This alters the visited states and prevents high sensitivity of policies to noiseless observations.
    \label{fig:roll_drop_system}}
\end{figure}

\section{Introduction} \label{sec:intro}
\Gls{drl} gained traction in the legged robotics community as a promising approach to the control problem, enabling highly dynamic and sophisticated locomotion capabilities~\citep{Lee2019, yang2020multi, kumar2021rma}. 
The sample complexity associated with high-dimensional problems such as locomotion, the risk of breaking the machines at the beginning of the training and the difficulty of resetting the robots make the use of physics simulators~\citep{raisim, makoviychuk2021isaac} appealing for training \gls{drl} control policies. 
However, this convenience often requires addressing the \textit{reality gap} between the simulated training and physical deployment domains. 

Strategies to address such a reality gap include accurately identifying properties such as \gls{com}, mass and inertia of robot links, impedance gains, system communication delays, friction, and actuation dynamics~\citep{Hwangbo2019, Lee2020}.
In addition, relevant distributions suitable for domain randomisation need to be selected \citep{Tan2018, Lee2019}; as part of such randomisation of the environment, sensory noise needs modelling and it is introduced in simulation during  training \citep{Jakobi1995, Hwangbo2019}.

We recently proposed \gls{erfi} \citep{campanaro2022} to handle system and actuation uncertainty as an alternative to a complete system and distribution identification for dynamics randomisation.
We demonstrate state-of-the-art sim-to-real performances by only randomising (and tuning) two parameters.
However, the robustness showed by \gls{erfi} in challenging conditions did not explicitly encompass modelling noise in observations.

In this work, we propose Roll-Drop, a method that improves the robustness of \gls{drl}-based locomotion controllers to observation noise by introducing dropout during rollout.
In continuation with \gls{erfi}'s simplicity, Roll-Drop only needs tuning a single parameter.

In the following sections, we present the method, analyse the results, and compare the robustness of alternatives to the injection of noise in the state space of the policy. 
Roll-Drop demonstrates an 80\% success rate when up to 25\% noise is injected in the observations, whereas in the same conditions other techniques experienced less than 40\% success rate.
The policies were trained in simulation on flat ground and deployed on a Unitree A1.


\section{Related Work} \label{sec:related_works}
Modern robots are equipped with diverse sensors to ensure acceptable levels of autonomy by estimating either the robot's state or the surrounding environment.
Such sensors include \glspl{imu}, joint encoders \citep{Hubicki2016}, torque and contact sensors \citep{Hutter2016}, RGBD cameras \citep{Rudin2021, Gangapurwala2022, Miki2022}, and lidar scanners \citep{Mattamala2022}.
\Gls{drl} approaches applied to locomotion controllers conveniently train policies that can take advantage of such rich sensory information.

Simulators are paramount here to reducing costs and training time while ensuring safety during the delicate training procedure.
Moreover, simulators provide the repeatability necessary to investigate eventual undesired behaviour.
However, in contrast to real sensors, simulators provide perfect and noiseless information far from what the policy would experience when deployed on a real robot, causing an additional sim-to-real gap to be addressed.

Research has focused on tackling the mismatch between simulated and real sensors by directly modelling the noise from real systems and injecting it into the network state during training.
\cite{Hwangbo2019} sample the joint velocity noise from uniform distributions, similarly to \cite{Lee2020} for linear and angular velocity noise; these were then added to the simulator's observations to improve robustness.
\cite{Bohez2022} use instead normal distributions to model observation noise for joint positions, angular velocity, linear acceleration, and base orientation, while \cite{Siekmann2020, Yu2022} also include the joint-encoder offsets, which were sampled from a uniform distribution. 
\cite{Gangapurwala2022, Miki2022}, instead, focus on exteroceptive sensors and inject noise into the height maps to foster the controller robustness to artefacts and sudden spikes.

Additionally, determining the noise characteristics is a delicate and costly process.
Often little detail on the process is provided, and ablation studies supporting the necessity of such randomisation are absent.
Roll-Drop addresses this lack of information as one parameter is enough to characterise the implementation.

\section{Problem Definition: Blind Quadrupedal Locomotion}
We model a quadrupedal system as a floating base $B$ described by the reference frame $\mathcal{B}$, represented w.r.t. a world reference frame $\mathcal{W}$, whose $z$-axis aligns with the gravity axis.
$\mathcal{B}$'s $x$-axis $x_\mathcal{B}$ points in the forward direction of motion of $B$, the $y$-axis $y_\mathcal{B}$ to the left and the $z$-axis $z_\mathcal{B}$ upwards.
The base position is then  expressed as  $r_{B}\in\mathbb{R}^3$, and the orientation, $\mathrm{q}_{B}\in\mathit{SO}(3)$, is represented by a unit quaternion, whose corresponding rotation matrix is denoted as $\mathbf{R}_{B}\in\mathit{SO}(3)$.

In this work, we will employ a Unitree A1 quadruped, whose four legs are composed of three joints each.
We will refer to the front-right leg as \texttt{FR}, to the front-left leg as \texttt{FL}, to the hind-right as \texttt{HR}, and to the hind-left leg as \texttt{HL}.
Each leg has a hip adduction/abduction \texttt{HAA}, hip flexion/extension \texttt{HFE}, and knee flexion/extension \texttt{KFE} joint.
For example, we refer to the front-right hip flexion/extension as \texttt{FR\_HFE}.
The vector $\mathrm{q}_{j}\in\mathbb{R}^{n_{j}}$ -- in our system, $n_j=12$ -- contains the angular positions of the rotational joints of all limbs, which are actuated through an impedance control, simplified as described by \citet{peng2017learning}:
\begin{equation}
    \Gamma_j = K_p(\mathrm{q}^{\ast}_j-\mathrm{q}_j)
    - K_d\mathrm{\dot{q}}_j.
\label{eq:impedance_controller_simplified}
\end{equation}
where $\Gamma_j$ are the actuation torques on the joints, $\mathrm{q}^{\ast}_j$ is the vector representing desired joint positions, and $K_p$ and $K_d$ refer to the position and  velocity tracking gains, respectively, which in our system are $K_p = 15.0$ and $K_d = 1.0$.

\subsection{Reinforcement Learning}
The \gls{rl} problem is modelled as an \gls{mdp} including a state space $\mathbf{S}$, an action space $\mathbf{A}$, an initial state distribution $p_1(s_1)$, a transition dynamics $p(s_{t+1}|s_t,a_t)$ compliant with the Markov property $p(s_{t+1}|s_1,a_1,\dots, s_t,a_t) = p(s_{t+1}|s_t,a_t)$ for any trajectory $\Tau_{1:t} = [(s_1,a_1,r_1),(s_2,a_2,r_2),\dots,(s_t,a_t,r_t)]$, where $r_i = R(s_i, a_i)$ is the reward obtained from a reward function $R: \mathbf{S}\times\mathbf{A}\rightarrow\mathbb{R}$.
In all the previous, $s_i \in \textbf{S}$ and $a_i \in \textbf{A}$.

A policy -- in our case, the controller -- selects actions in the \gls{mdp} given a specific state.
The policy -- denoted by $\pi_\theta$, where $\theta \in \mathbb{R}^n$ is a vector of $n$ parameters -- is stochastic, and $\pi_\theta(a_t|s_t)$ is the conditional probability density of $a_t$ associated with the policy.
The agent uses its policy to interact with the \gls{mdp}, realising the trajectory of states, actions, and rewards $\Tau_{1:T} = (s_1,a_1,r_1),\dots,(s_T,a_T,r_T)$.

The policy $\pi$ is trained through an optimisation problem to maximise the cumulative discounted reward it obtained from the starting state, expressed as $\pi^*=\textrm{argmax}\:\mathbb{E}[r_1^\gamma|\pi]$, where $r_t^\gamma$ is the total discounted reward from time-step $t$ onward, as $r_t^\gamma = \sum_{k=t}^{T}\gamma^{k-t}r(s_k,a_k)$, where $0<\gamma<1$.

\subsection{Implementation}
The quadruped robot is required to follow a velocity command $s_c = [v_x, v_y, \dot{\psi}]_\mathcal{B}$ on flat ground using proprioceptive information.
Here $v_x$ and $v_y$ are the linear velocities along $x_\mathcal{B}$ and $y_\mathcal{B}$ respectively, while $\dot{\psi}$ is the angular velocity around $z_\mathcal{B}$.

The state is represented as $s := \langle s_r, s_v,  s_{j_p}, s_{j_v}, s_a, s_f, s_c \rangle \in {\mathbb{R}}^{196}$, where $s_r^\mathcal{B} \in {\mathbb{R}}^{3}$ is the last row of the rotation matrix $\mathbf{R}_{B}$, $s_v \in {\mathbb{R}}^{6}$ is the base linear and angular velocities, 
$s_{j_p}^\mathcal{B} \in {\mathbb{R}}^{84}$ is the history of joint position errors and $s_{j_v}^\mathcal{B} \in {\mathbb{R}}^{84}$ is the history of joint velocities, $s_a \in {\mathbb{R}}^{12}$ is the previous action, $s_f \in {\mathbb{R}}^{4}$ is the contact state of the feet, and $s_c^\mathcal{B} \in {\mathbb{R}}^{3}$ is the velocity command. 
The actions $a \in {\mathbb{R}}^{12}$ are retrieved from the policy $\pi$ -- implemented as a \gls{mlp} formed by three layers of size $[512, 256, 256]$ -- and interpreted as the reference joint positions $\mathrm{q}^*_j$, tracked by the impedance controller in \Cref{eq:impedance_controller_simplified}. 
The onboard state estimator does not provide the base linear velocity in $s_v$; hence, we estimate it and $s_f$ similarly to \cite{Ji2022} through an \gls{mlp} of size $[128, 128, 128]$.
We train $\pi$ on flat ground using \gls{ppo} \citep{Schulman2017} until convergence (\Cref{fig:reward}), adopting the rewards and hyper-parameters in \Cref{tab:hyper_params_a1_ppo}.

\begin{table}
\caption{(a) \Gls{ppo} hyper-parameters used for training the Unitree A1 policy; (b) the rewards adopted during training, and their weights.}
\label{tab:hyper_params_and_rewards}
\centering
\small
\subtable[]{
\begin{tabular}{l|l}
    Hyperparameter & Value \\
      \midrule
     Control dt & 0.02 [s]\\
     Sim dt & 0.002 [s]\\
     Batch size & 25600\\
     Mini-batch size & 6400\\
     Number of epochs & 8\\
     Clip range & 0.2\\
     Entropy coefficient & 0.\\
     Discount factor & 0.996\\
     GAE discount factor & 0.95\\
     Learning rate & $1e^{-4}$\\
\end{tabular}
\label{tab:hyper_params_a1_ppo}
}
\hfill
\subtable[]{
\begin{tabular}{l|lr}
      & Definition & Weight\\
      \midrule
     Base orientation & $k_c\cdot||\mathbf{R}_{B}^z - [0, 0, 1]||^2$ & $-30$ \\
     Base linear velocity & $\phi(v_{b_{x, y}}^*, v_{b_{x, y}}, 5)$ & $15$\\
     Base angular velocity & $\phi(v_{b_{z}}^*, \omega_{b_{z}}, 5)$ & $15$ \\
     Action smoothness & $k_c\cdot||\mathrm{q}^{\ast}_{j_{t}} - \mathrm{q}^{\ast}_{j_{t-1}}||^2$ & $-7$\\
     Feet clearance & $k_c\cdot\sum_{n=0}^{n<4} (0.1 - f_{z_n})^2$ & $-400$ \\
     Feet sleep & $k_c\cdot||\dot{f}_{x, y}||^2$ & $-8$\\
     Joint position & $k_c\cdot||q_j - \mathrm{q}_j^N||^2$ & $-4$ \\
     Joint velocity & $k_c\cdot||\dot{q_j}||^2$ & $-0.01$ \\
     Joint torque & $k_c\cdot||\tau_j||^2$ & $-0.4$ \\
     Feet swing duration & $\sum_{n=0}^{3}(\mathbf{t}_{air, n} - 0.5)$ & $8$\\
     Pronking gait & $k_c\cdot\sum_{n=0}^{3}(f_{c_n}\cdot1)$ & $-35$\\
\end{tabular}
\label{tab:rewards}
}
\end{table}


\section{Roll-Drop} \label{sec:roll_drop}

The proposed method, Roll-Drop, exploits the concept of using dropout to mimic an observation noise to improve the network's robustness in a sim-to-real deployment scenario.
In particular, {Roll-Drop} adds a customised dropout layer \citep{hinton2012}, active only during rollouts and turned off during training.
The resulting random perturbations (as shown in \Cref{fig:roll_drop_system}) cause the policy $\pi$ to explore regions of the state space $\textbf{s}$ and action space $\textbf{A}$ different from the standard training, as in \Cref{fig:drop_distr}.

%



When the dropout is not present, the actions are sampled with a policy $\pi_\theta(a|s)$: $\pi_\theta(a|s)$ = $\mu_\theta(s) + \mathcal{N}(0,\sigma)$, where $\mu_\theta(s)$ is the output of the network.
When, instead, dropout is included the parameters $\theta$ become $\hat{\theta}=\theta + \delta\theta$, and consequently $\pi_{\theta+\delta\theta}(a|s)$ = $\mu_{\theta+\delta\theta}(s) + \mathcal{N}(0,\sigma)$.
Based on this, in a state $s$: $\mu_\theta(s)\mapsto\alpha$, while $\mu_{\theta+\delta\theta}(s)\mapsto\hat{\alpha}$ with $\hat{\alpha}=\alpha+\delta\alpha$ and $\delta a$ function of $\hat{\theta}$.

Assuming deterministic dynamics and same initialisation, the transition probability can be reformulated as the transition function $\mathcal{P}(s,\alpha)\rightarrow\zeta$, when dropout is inactive, and $\mathcal{P}(s,\hat{\alpha})\rightarrow\hat{\zeta}$ otherwise, where $\zeta$ and $\hat{\zeta}$ are the next states. 
Similarly to $\hat{\alpha}$, $\hat{\zeta}$ can be expressed as $\hat{\zeta}=\zeta+\delta\zeta$, where $\delta\zeta = f(\alpha+\delta\alpha)$.

At the next time-step ($t+1$), when dropout is inactive we can expect $\mathcal{P}(s',a')\rightarrow s''$, whereas when dropout is active $\mathcal{P}(s'+\delta s',a'+\delta a')\rightarrow s''+\delta s''$.
Here $\delta s'$ and $\delta s''$ represent the discrepancy between the transitions happening adopting $\pi_\theta(a|s)$ and $\pi_{\theta+\delta\theta}(a|s)$.

In this work, we added a single layer of Roll-Drop after the second layer of the \gls{mlp} network.
Notably, since the dropout-injected noise happens only during rollout, $\pi$ develops reflexes to recover from dangerous states and becomes more robust to perturbations; conversely, adding dropouts during training does not allow the policy to develop reactions to perturbations.


\subsection{Tuning Roll-Drop probability} \label{sec:tuning-roll_drop}
Similarly to other randomisation techniques \citep{Tobin2017, Valassakis2020, campanaro2022}, the tuning of the Roll-Drop probability is carried out empirically: At first, the environment (defined in \Cref{tab:hyper_params_a1_ppo,tab:rewards}) is tuned for tracking a velocity command $s_c$ on flat ground \textbf{without} any randomisation and using a fixed random seed.
After the policy converges to the desired behaviour, the Roll-Drop layer is included in the network, and the dropout probability is increased (starting from $p=0.$) until the training is stable again.
This can be seen in \Cref{fig:dropout_inference}, where we tested different dropout probabilities and how they affected the training convergence.

 \begin{figure}
  \centering
  \hfill
  \subfigure[]{
    \centering
    \includegraphics[width=0.32\linewidth]{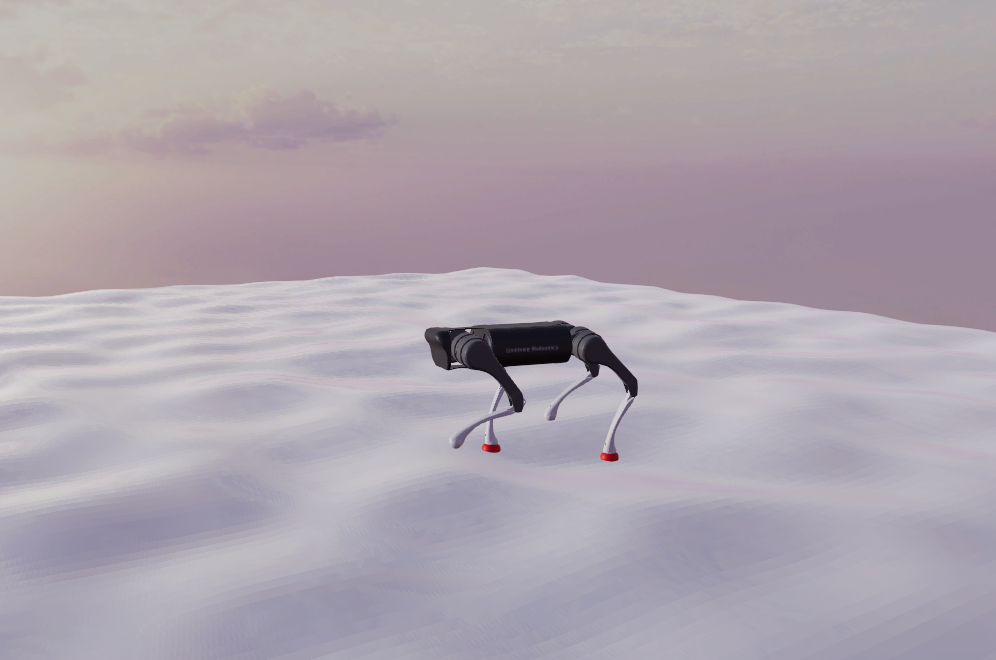}
       \label{fig:rough_terrain}
    }
\hfill
  \subfigure[]{
  \small
  \begin{tabular}{l|l}
    Simulator parameter & Value\\
    \midrule
     $K_p$ & 15\\
     $K_d$ & 1\\
     Torque Limit & 20 [N/m]\\
     Motor Act. Delay & 0.012 [s]\\
     Motor Static Friction & 0.2\\
     Motor Dyn. Friction & 0.01\\
     Ground Friction & 0.4\\
     Gravity & -9.81 $[m/s^2]$\\
    \end{tabular}
    \label{tab:default_settings_experiments}
  }
  \hfill~
  \caption{
    (a) Evaluation environment: the non-flat terrain is more realistic and it brings stochasticity to the evaluation of the robustness to observation noise.
    (b) Default settings for the testing environment.}
   \vspace{-0.4cm}
\end{figure}

\section{Experimental Setup} \label{sec:experimental_design}
To assess the performance of the method proposed we run several experiments with different levels of noise affecting the observations.
The environment's settings are fixed as in \Cref{tab:default_settings_experiments}, the robot is commanded a constant velocity $s_c = [v_x, v_y, v_z]$, where $v_x=0.5 [m/s]$ is the only non zero component.

Alongside these settings we included a mild rough terrain to better represent realistic conditions, as in \Cref{fig:rough_terrain}.
In the environment defined as above we varied the amount of noise ($n$) from 0\% to 60\% as in \Cref{eq:noise_model}, with a step of 5\%, and 100 experiments were run for each noise configuration (randomising the spawning point of the robot on the rough terrain).
\begin{equation}
    s_t \leftarrow s_t + n \cdot \mathcal{U}(-1, 1) \cdot s_t \textrm{, where n} \in [0, 0.6)
\label{eq:noise_model}
\end{equation}


The success rate in \Cref{fig:dropout_success_rate} is measured across the 100 experiments carried out for each percentage of injected noise.
To successfully complete the evaluation the robot does not have to fall on the ground and it has to walk for at least 1 [m] in the direction of the velocity commanded, if one of the two conditions is not respected the experiment is considered a failure.
The ratio between the successful runs and the total number runs gives the success rate.

\section{Results and Discussion} \label{sec:results}
We compared Roll-Drop ($p=0.0001$) against \textit{No Randomisation} --which is based on the original environment used for Roll-Drop but without dropout, in \Cref{sec:tuning-roll_drop}--, against \gls{erfi} \citep{campanaro2022} that demonstrated state-of-the-art robustness to external perturbations, against dropout during training ($p=0.001$), and finally a mixture of dropout during training ($p=0.001$) plus dropout during rollout ($p=0.0001$).
From the results in \Cref{fig:dropout_success_rate}, the most robust method to the injection of noise in the observations is Roll-Drop, which retained 80\% success rate when more than 25\% of noise was injected.
The performance of the policies trained with other techniques degrades quickly as soon as noise is injected, suggesting strong sensitivity to observation distribution encountered during training. Note that all the controllers were trained and tested adopting the same random seed.

\subsection{Dropout during training}
\begin{figure} [h!]
    \subfigure[]{
      \centering
      \includegraphics[width=0.48\textwidth]{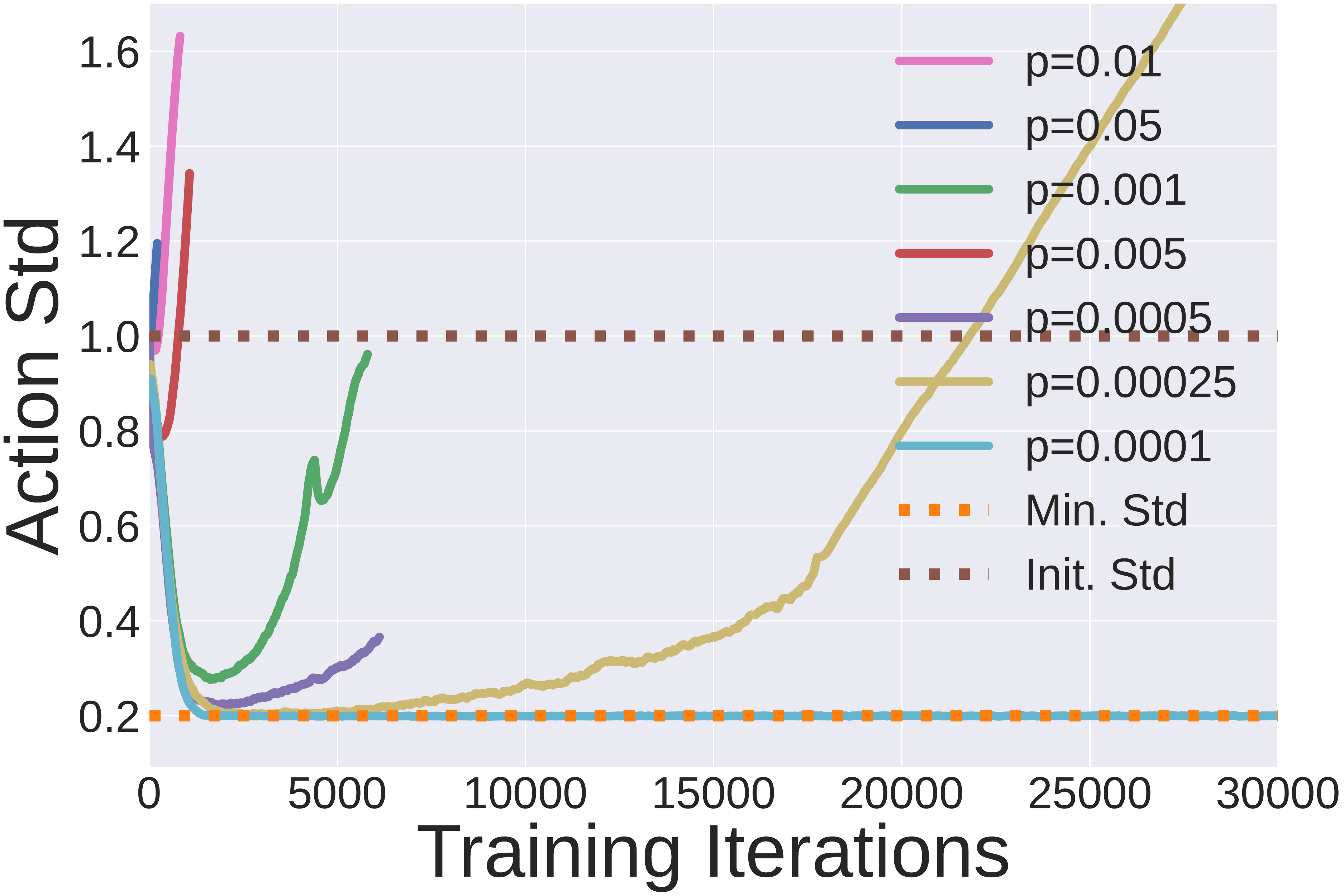}
      \label{fig:dropout_inference}
    }
    \subfigure[]{
      \centering
      \includegraphics[width=0.48\textwidth]{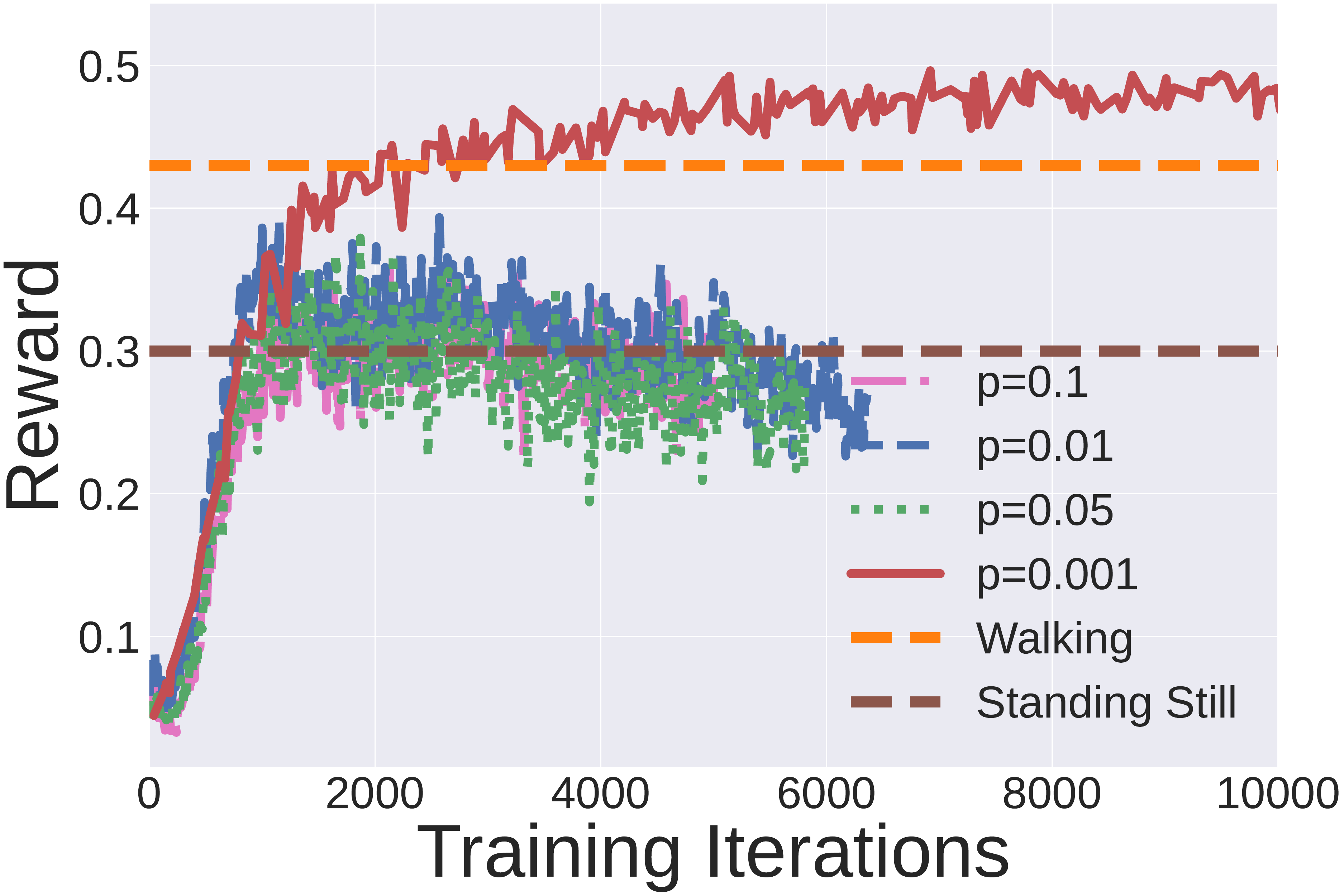}
      \label{fig:dropout_training}
    }
    
    \caption{
    In \Cref{fig:dropout_inference} the introduction of dropout during rollouts heavily affects the convergence, in this figure we show the effect of different dropout probabilities on the standard deviation used to sample actions in policy gradient algorithms.
    In \Cref{fig:action_distr} alongside dropout during rollouts, we investigate the effects of dropout during training.
    When the dropout probability is too high the policy displays a standing-still behaviour, which suggests its inability to correlate inputs to outputs.
    }
\end{figure}

We motivate the adoption of dropouts during rollouts (Roll-Drop) in \Cref{sec:roll_drop}, nonetheless we investigated the performances resulting from adopting dropouts during training, and during both training and rollouts.
This is depicted in \Cref{fig:dropout_training}, where we show the effects of different dropout probabilities when it is applied during training and not during rollout.
A cluster of lines can be identified with dropout probability $\in [0.01, 0.1]$, their maximum reward oscillates around 0.3, which corresponds to the robot standing still.
Based on our experience, the randomness injected by dropout does not allow the network to correlate  inputs and outputs well, and by standing still the policy avoids the termination reward (when the robot falls on the ground), while still receiving some positive points from the rewards in \Cref{tab:rewards}.
Indeed, as soon as the dropout probability is lowered to $p=0.001$  the total reward increases, and the robot starts walking again.
We compare the performance of the policies trained 1) with only dropout during training ($p=0.001$), 2) with the dropout during training ($p=0.001$) plus dropout during rollout ($p=0.0001$), 3) with the policy trained with Roll-Drop only.
In \Cref{fig:dropout_success_rate}, we can observe that the introduction of dropout during training, even in conjunction with dropout during rollouts is detrimental.

\subsection{Dropout probability and convergence}
The classical usage of dropout in supervised learning is to regularise the learning of the employed networks~\cite{srivastava2014droput}.
Randomly dropping units from the \gls{nn} during training prevents them from co-adapting, thus significantly reducing overfitting; at test time, then, the dropout is removed to approximate averaging the predictions of all these partial networks by using a complete network with smaller weights.
A second application of dropout is to approximate a Bayesian network~\cite{gal2016dropout}: in this case, applying dropout at inference time, we can generate multiple predictions feeding the network multiple times with the same input.
This gives us a probability distribution of the outputs which we can then analyse. 
For such applications dropout probability typically varies between 20\% and 50\%, but can reach values up to 80\%~\cite{srivastava2014droput}.

As can be seen from \Cref{fig:dropout_inference}, in the case of Roll-Drop the probability is much lower: 0.01\%.
In fact, differently from supervised/semi-supervised/non-supervised learning where some kind of target is provided, for \gls{rl} the policy loss has a moving target dependent on the actions taken by the current policy itself.
The better states the policy explores, the higher the reward it will receive, conversely exploring bad states can result in the policy exploring a wider actions space, and eventually to catastrophically diverging to even worse states.
Moreover, when some neurons are dropped during rollouts the noise introduced affects the following state of the episode.
Considering the latter in conjunction with having a moving target, it is clear that \gls{rl} is more sensitive to dropout probabilities, and that lower dropout probabilities are expected.

We can observe, in \Cref{fig:dropout_inference}, how for dropout probabilities $\in [0.00025, 0.01]$ the training diverges, while for $p=0.0001$ it converges to a stable behaviour. 
In fact, the way policy gradient algorithms like \gls{ppo} \citep{Schulman2017} explore the action space is based on sampling from a distribution $\mathcal{N}(\mu, \sigma)$, where $\mu$ is the action $a$ output of $\pi$, while $\sigma$ is a learnt parameter.
In \Cref{fig:dropout_inference} we plot the $\sigma$ for policies trained with different levels of dropout during rollouts (same random seed), the initial $\sigma_0=1.$, and it is capped to $\sigma_{min}=0.2$; high probabilities of dropout are responsible for the divergence.

\begin{figure}[ht!]
    \centering
    \includegraphics[width=\textwidth]{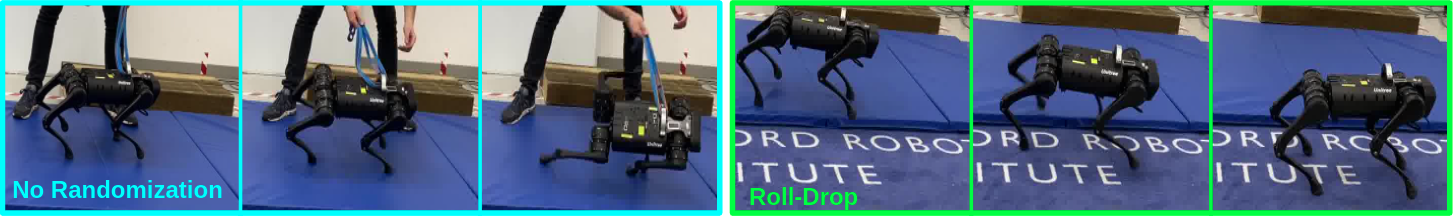}
    \caption{Policies trained with \textit{No randomisation} (left) and Roll-Drop (right), both trained on flat ground in simulation with $K_p=20$, and tested on the hardware with $K_p=15$. This change in gains highlights a case of system uncertainty, and demonstrates that Roll-Drop is capable of addressing sim-to-real gaps.
    \label{fig:hardware_experiment}}
\end{figure}

\subsection{How is Roll-Drop affecting the training?}
\begin{figure} [h!]
    \subfigure[]{
      \centering
      \includegraphics[height=3.5cm, width=0.31\textwidth]{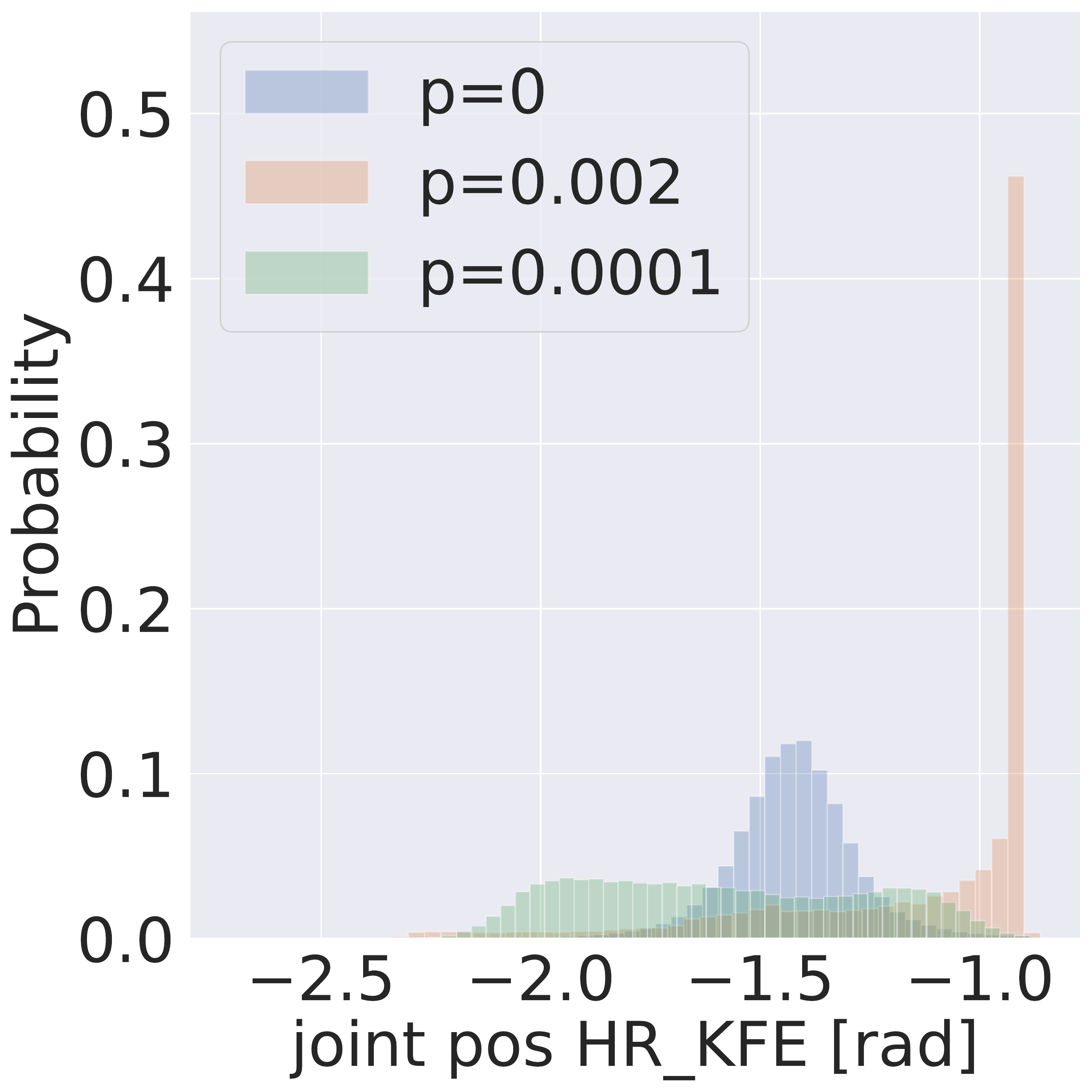}
      \label{fig:joint_pos_distr}
    }
    \subfigure[]{
      \centering
      \includegraphics[height=3.5cm, width=0.31\textwidth]{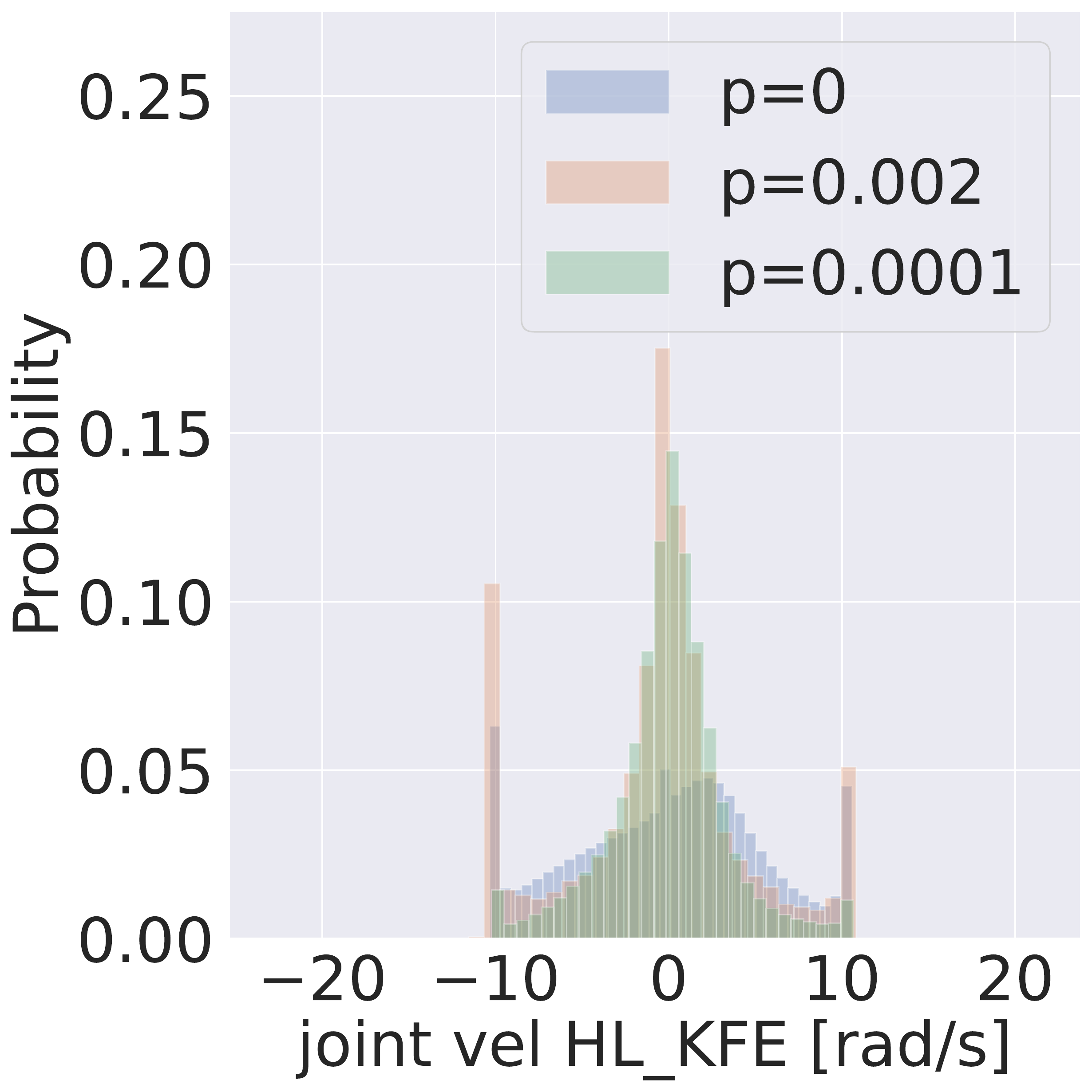}
      \label{fig:joint_vel_distr}
    }
    \subfigure[]{
      \centering
      \includegraphics[height=3.5cm, width=0.31\textwidth]{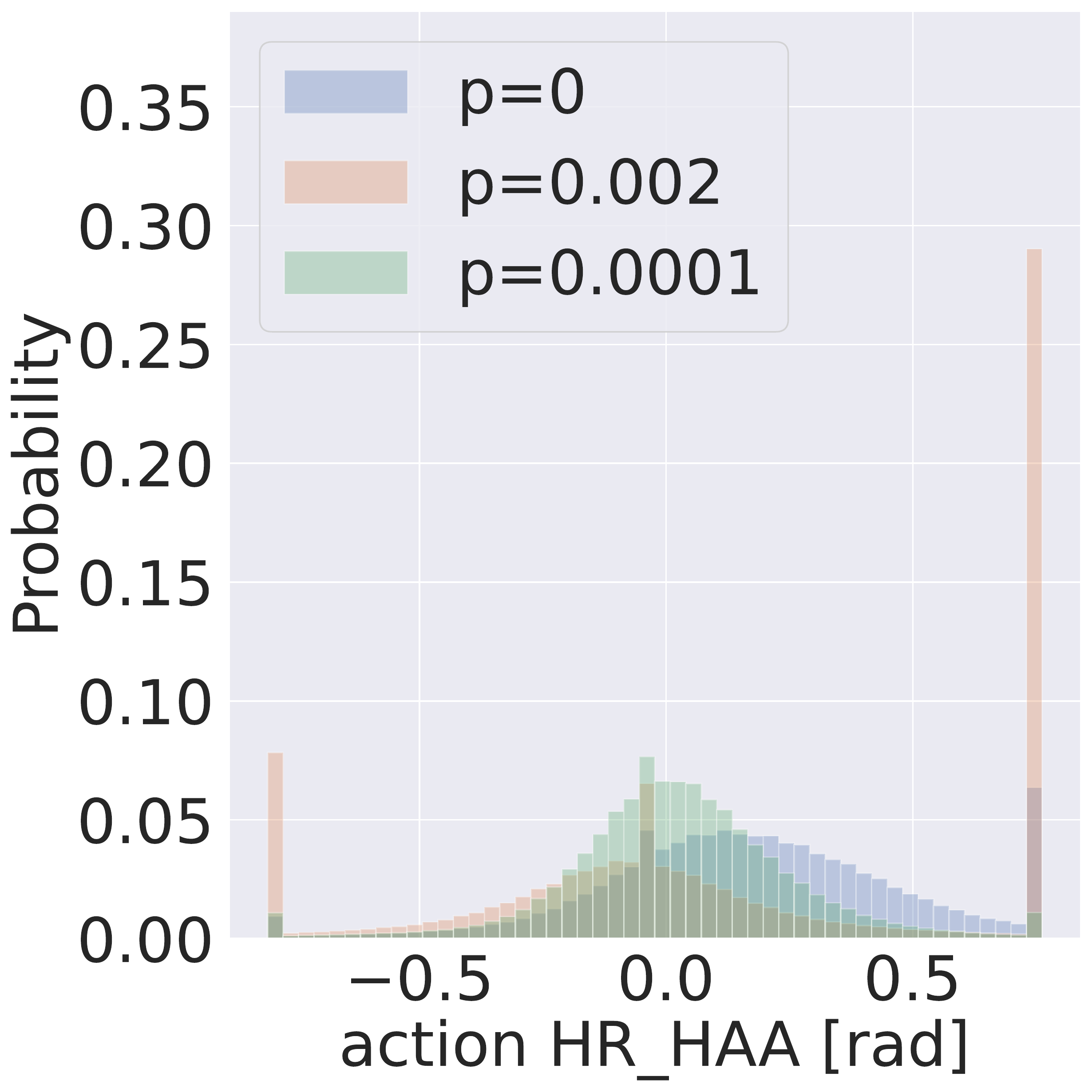}
      \label{fig:action_distr}
    }
    \subfigure[]{
      \centering
      \includegraphics[height=3.5cm, width=0.31\textwidth]{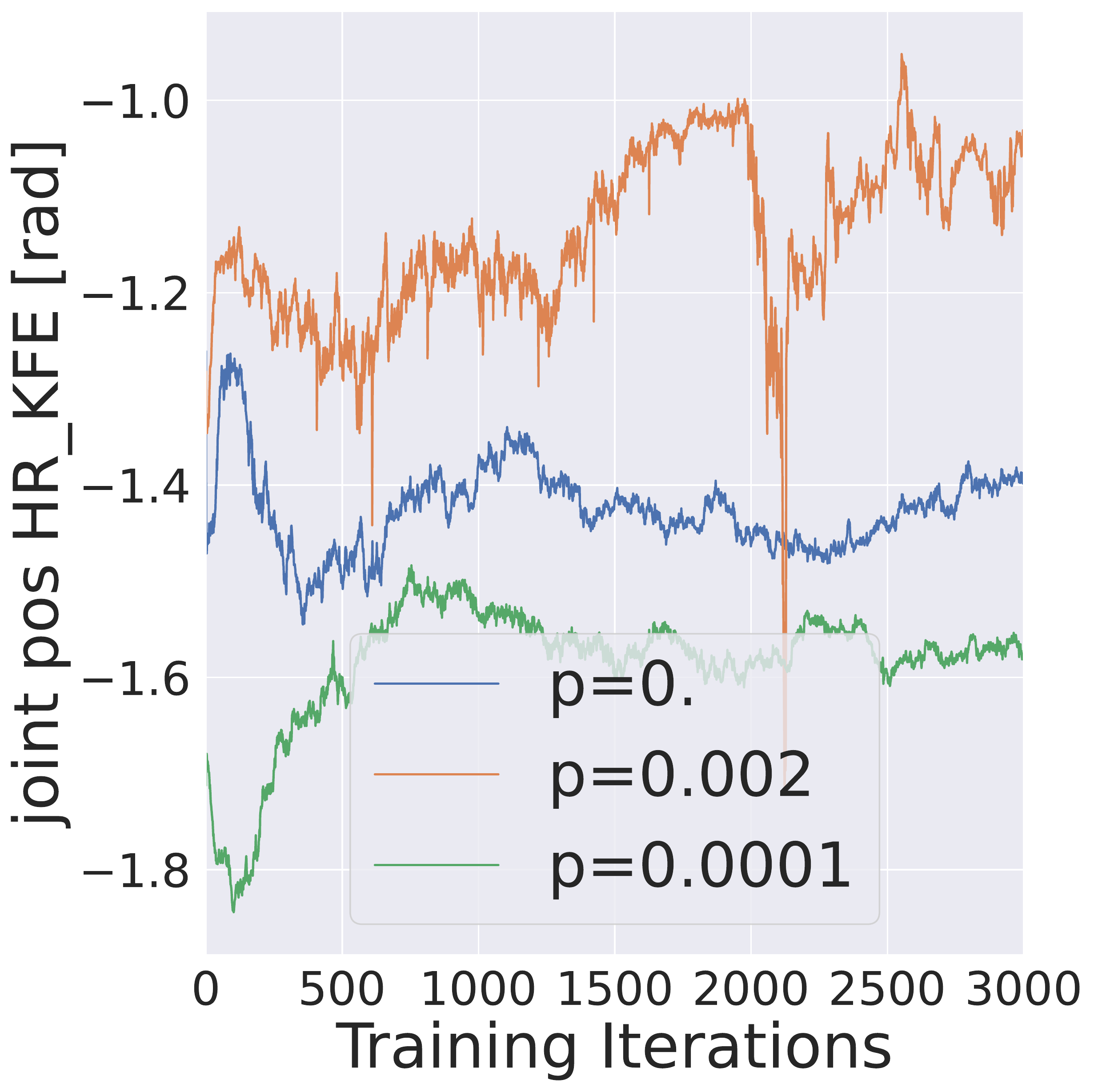}
      \label{fig:joint_pos_distr_running_mean}
    }
    \subfigure[]{
      \centering
      \includegraphics[height=3.5cm, width=0.31\textwidth]{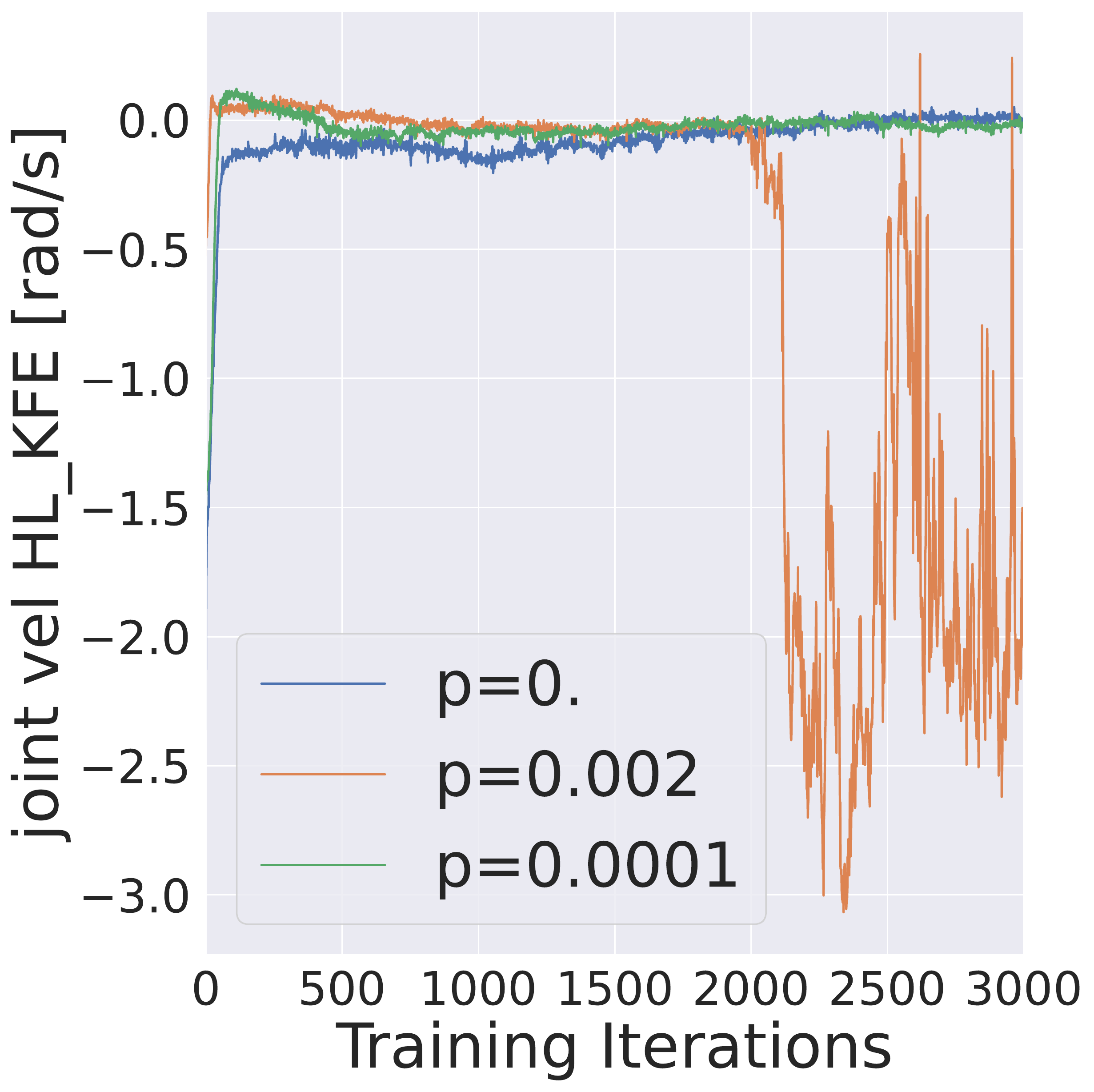}
      \label{fig:joint_vel_distr_running_mean}
    }
    \subfigure[]{
      \centering
      \includegraphics[height=3.5cm, width=0.31\textwidth]{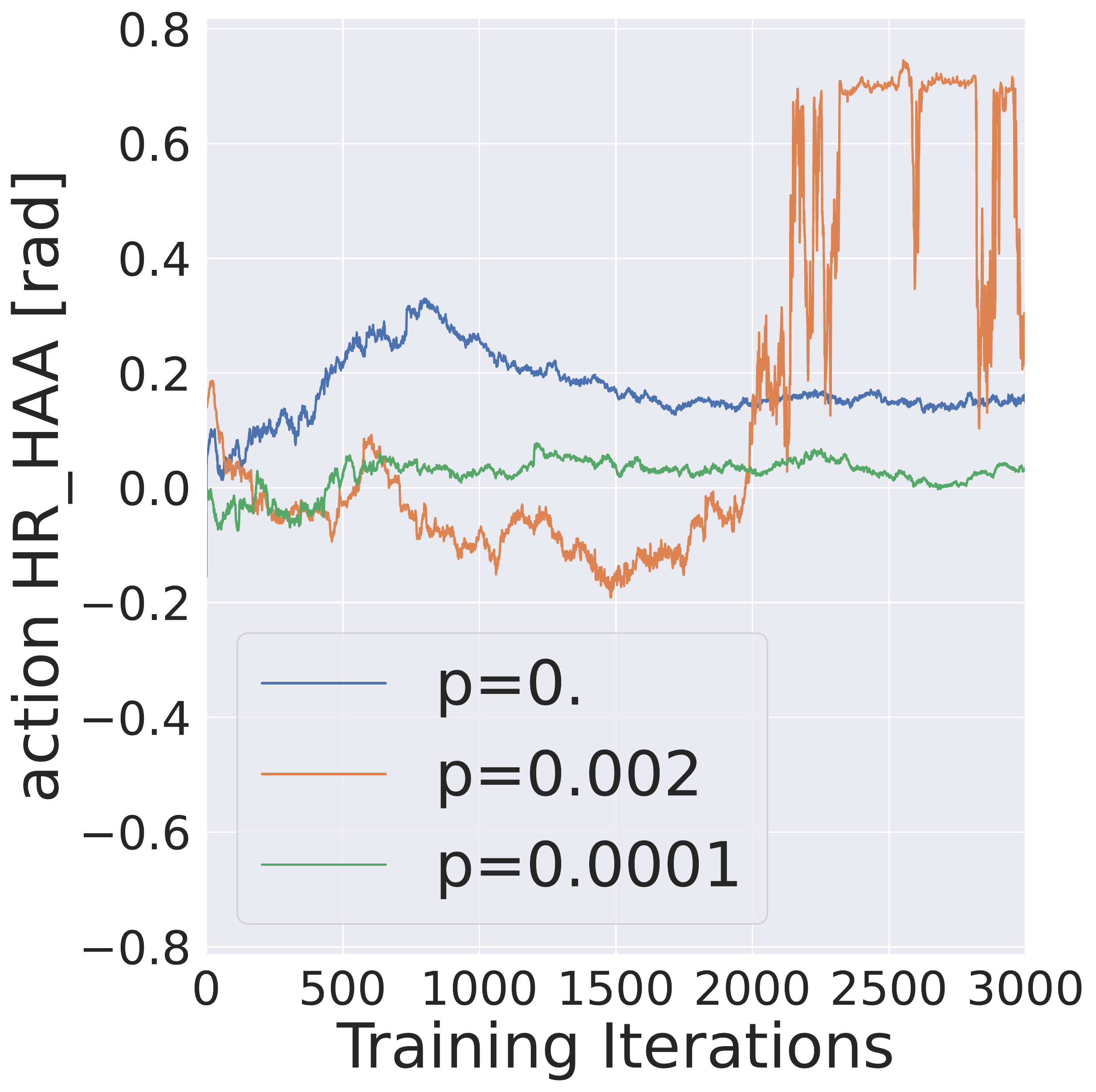}
      \label{fig:action_distr_running_mean}
    }

    \caption{
    \Cref{fig:joint_pos_distr,fig:joint_vel_distr,fig:action_distr} show some state/action distributions for Roll-Drop with three different probabilities: 0 (no Roll-Drop), 0.002, and 0.0001. When $p=0$ the policy is over-fitting to a fixed simulation environment, when $p=0.0001$ the policy converges to the desired behaviour, but exploring a different state/action space compared to $p=0$, and finally, when $p=0.002$ it diverges catastrophically and it explores bad portions of the state/action spaces (often the joint position/velocity limits) as can be seen from the orange histograms.
    Evidences of the divergence are provided in \Cref{fig:joint_pos_distr_running_mean,fig:joint_vel_distr_running_mean,fig:action_distr_running_mean}, which represent the mean for each state/action along the initial 3000 iterations considered. 
    }
    \label{fig:drop_distr} 
\end{figure}

In \Cref{fig:roll_drop_system}, we describe the effects of Roll-Drop on the training: the policies with and without dropout observe the same initial state (same random seed), and produce the same action until the first dropout is triggered.
After this event the two trainings take different trajectories $\Tau$ and $\Tau'$, because the policies output different actions and the robots end up in different states.
We investigated this further in \Cref{fig:drop_distr} by recording states and actions for the first 3000 training iterations (adopting 128 parallel environments and episodes of 4 [s]) for Roll-Drop with associated probabilities $\in [0., 0.002, 0.0001]$.
We show these distribution shifts for some states and actions: the joint position of \texttt{HR\_KFE} in \Cref{fig:joint_pos_distr}, the joint velocity of \texttt{HL\_KFE} in \Cref{fig:joint_vel_distr}, and the action of \texttt{HR\_HAA} in \Cref{fig:action_distr}.
These histograms demonstrate how such a tiny dropout probability --when compared to supervised/semi-supervised/non-supervised learning-- affects the training: For $p=0.0001$ the distributions of states and actions are different from $p=0.$, while for $p=0.002$ the training is clearly diverging (\Cref{fig:dropout_inference}) with most of the states and actions distributed close to the joint position and velocity limits.
Further evidences of the divergence are provided in \Cref{fig:joint_pos_distr_running_mean,fig:joint_vel_distr_running_mean,fig:action_distr_running_mean}, where we show the mean across all environments and time-steps for each of the first 3000 iterations.

\subsection{Different random seeds}
\begin{figure} [h!]
    \subfigure[]{
      \centering
      \includegraphics[height=0.34\textwidth]{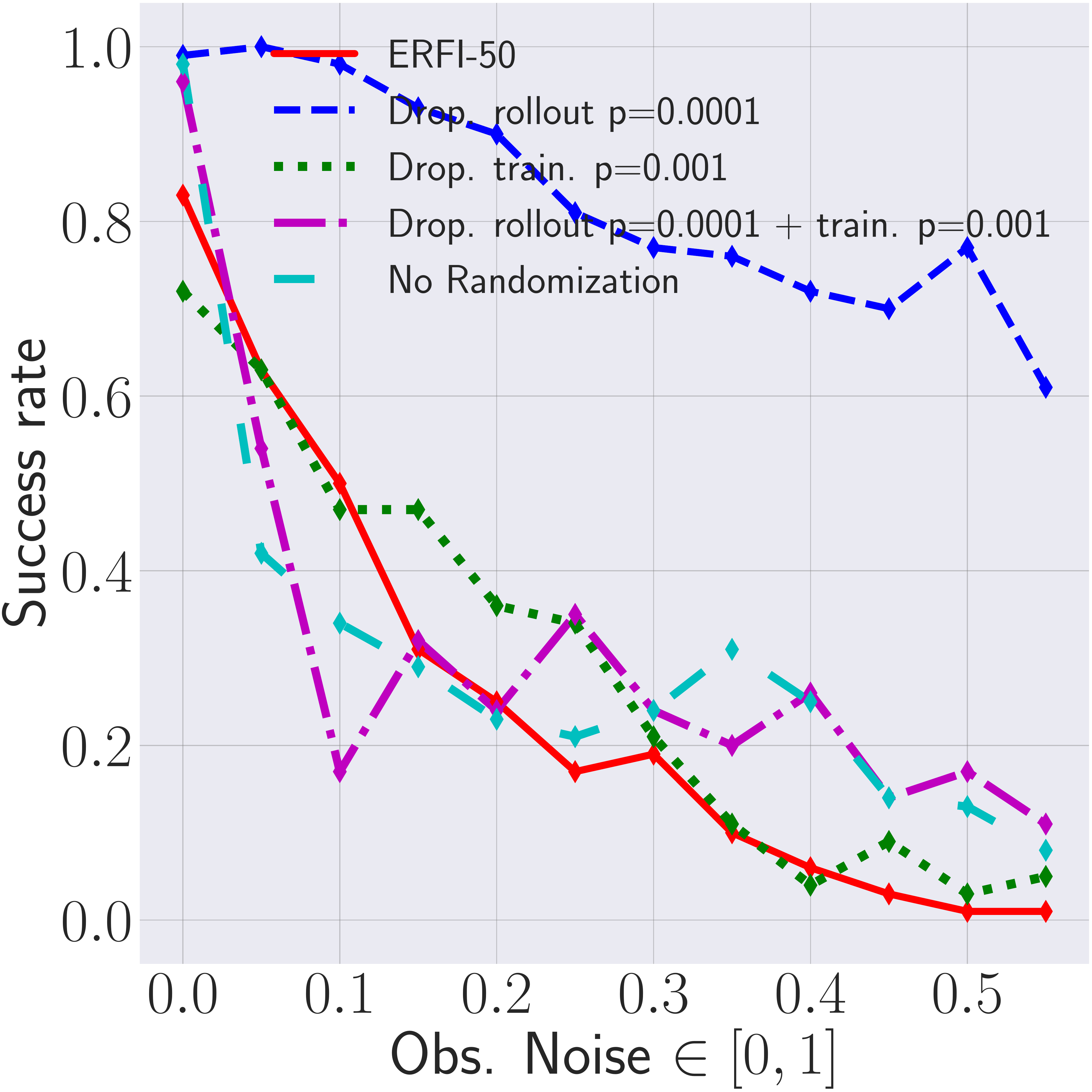}
      \label{fig:dropout_success_rate}
    }
    \subfigure[]{
      \centering
      \includegraphics[height=0.34\textwidth]{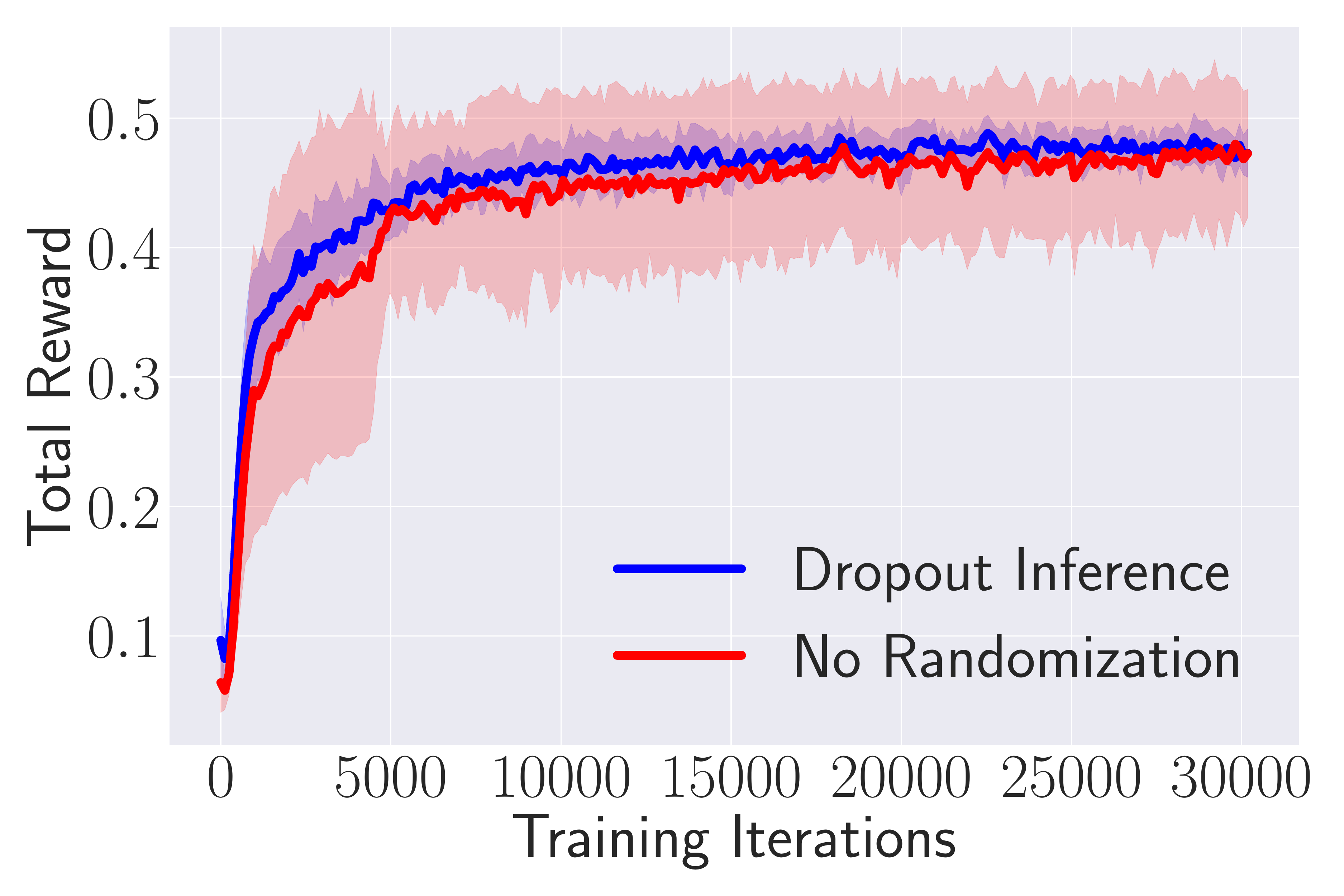}
      \label{fig:reward}
    }
    
    \caption{
    (a) Comparing the success rate of different methods in presence of noise in the observations.
    Roll-Drop performs more than twice as good as other methods, retaining 80\% of success rate when more than 25\% of noise was injected.
    (b) we trained the Roll-Drop and the \textit{No Randomisation} policies using 5 different random seeds, and here we compare their total rewards.
    Roll-Drop is more consistent, demonstrating in average higher total reward and a smaller standard deviation.
    Conversely, \textit{No Randomisation} is more sensitive to the random seed used: highest total reward on the seed on which the environment was tuned on, but the performance degraded when other seeds were adopted.
    }
\end{figure}

We investigated how consistent the training is when different random seeds are used: We trained five policies without any randomisation and five policies with Roll-Drop across five different seeds, and compared the total reward of both groups in terms of mean and standard deviation.
We considered the \textit{No Randomisation} setting as the perfect candidate for this comparison since it is massively over-fitting to the simulation environment.
From \Cref{fig:reward}, we discovered that Roll-Drop (blue line) is on average performing better, because of its higher robustness across different seeds, and this is also supported by the smaller standard deviation, when compared to \textit{No Randomisation}.
However, as expected, \textit{No Randomisation} is in absolute value performing better than Roll-Drop, but only for the seed the environment was originally tuned on; while for other seeds it gained lower rewards, and it has a more spread standard deviation.

\subsection{Training and deployment mismatch} \label{sec:training-and_deployment_mismatch}
\begin{figure} [h!]
    \subfigure[]{
      \centering
      \includegraphics[height=3.5cm,width=0.31\textwidth]{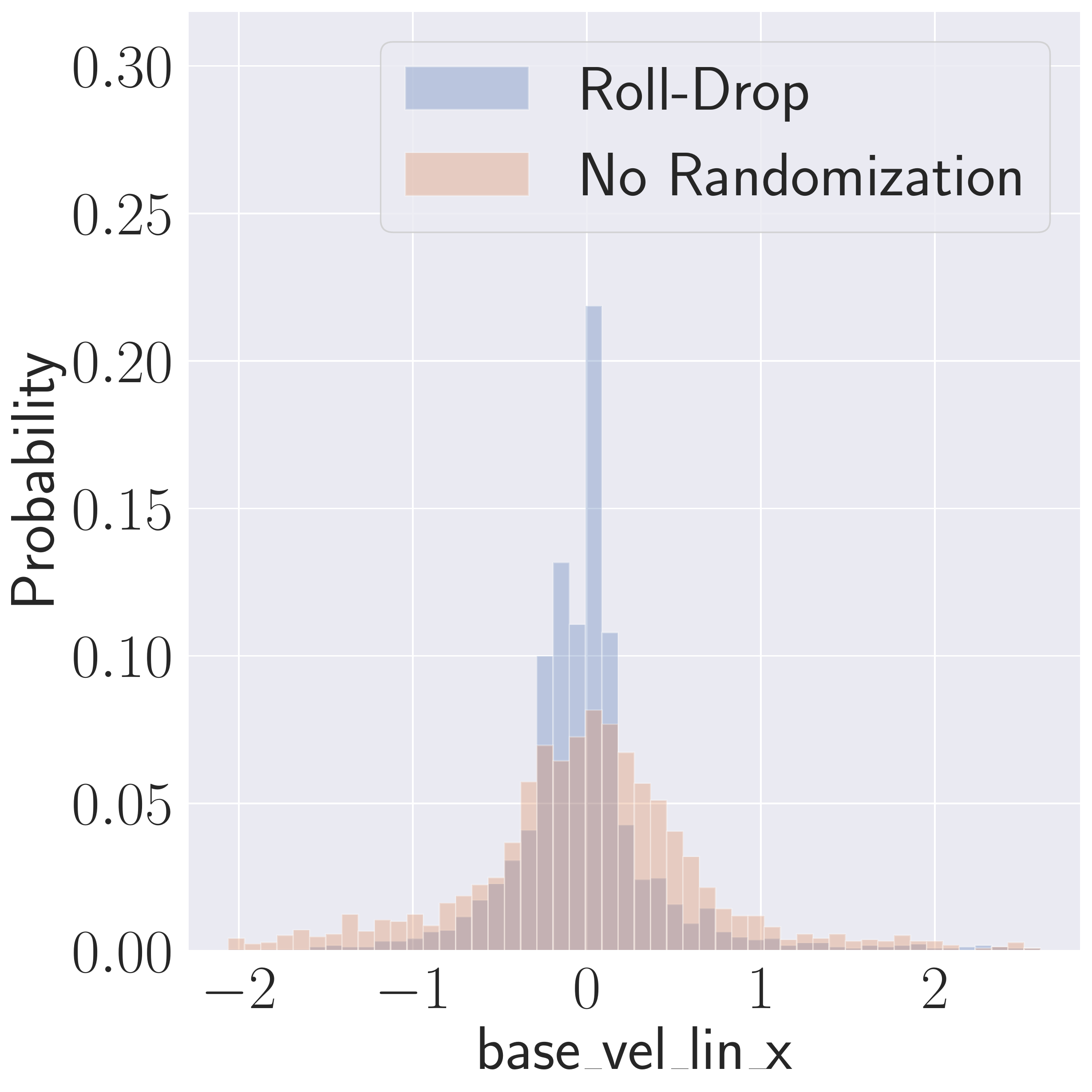}
      \label{fig:vel_lin_x}
    }
    \subfigure[]{
      \centering
      \includegraphics[height=3.5cm,width=0.31\textwidth]{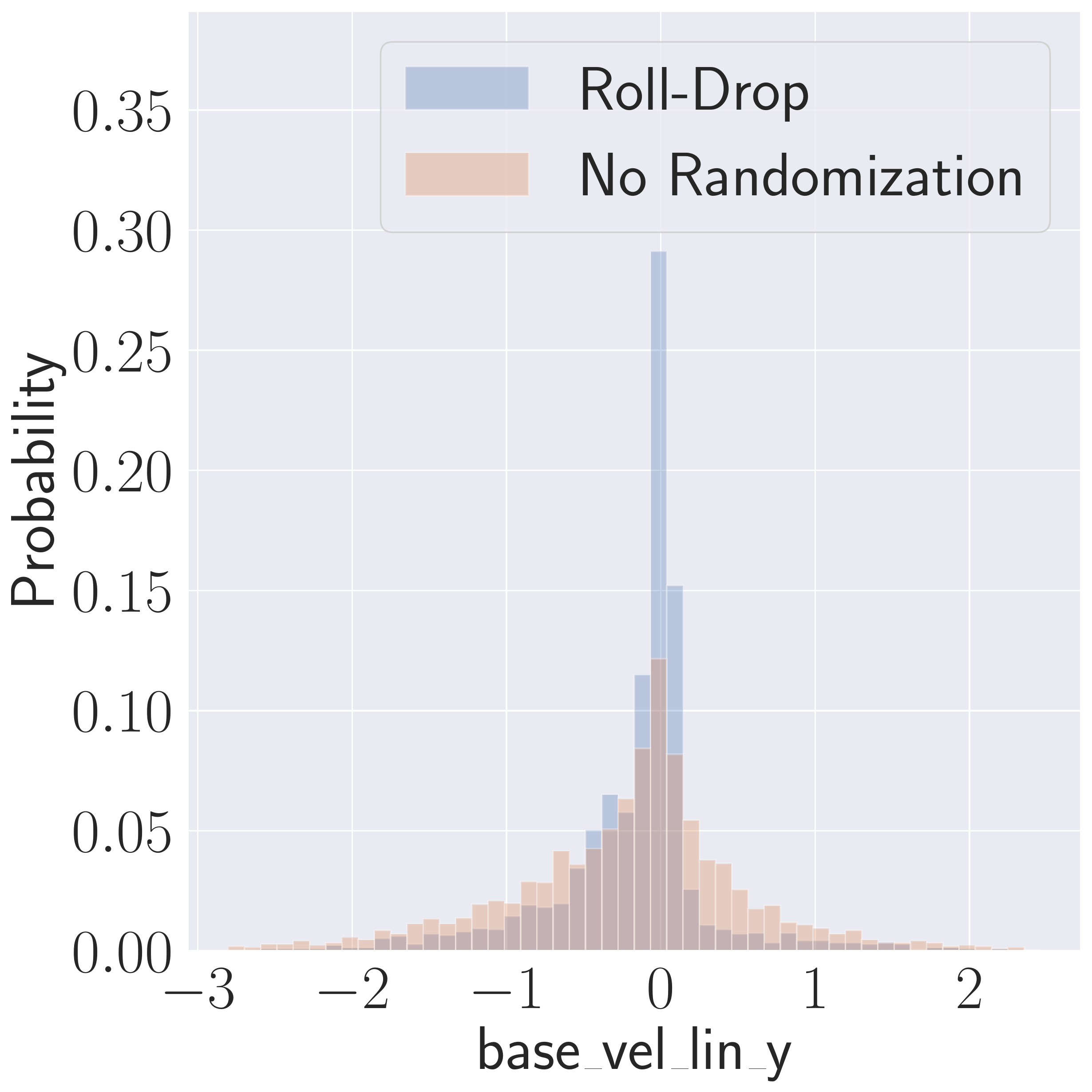}
      \label{fig:vel_lin_y}
    }
    \subfigure[]{
      \centering
      \includegraphics[height=3.5cm,width=0.31\textwidth]{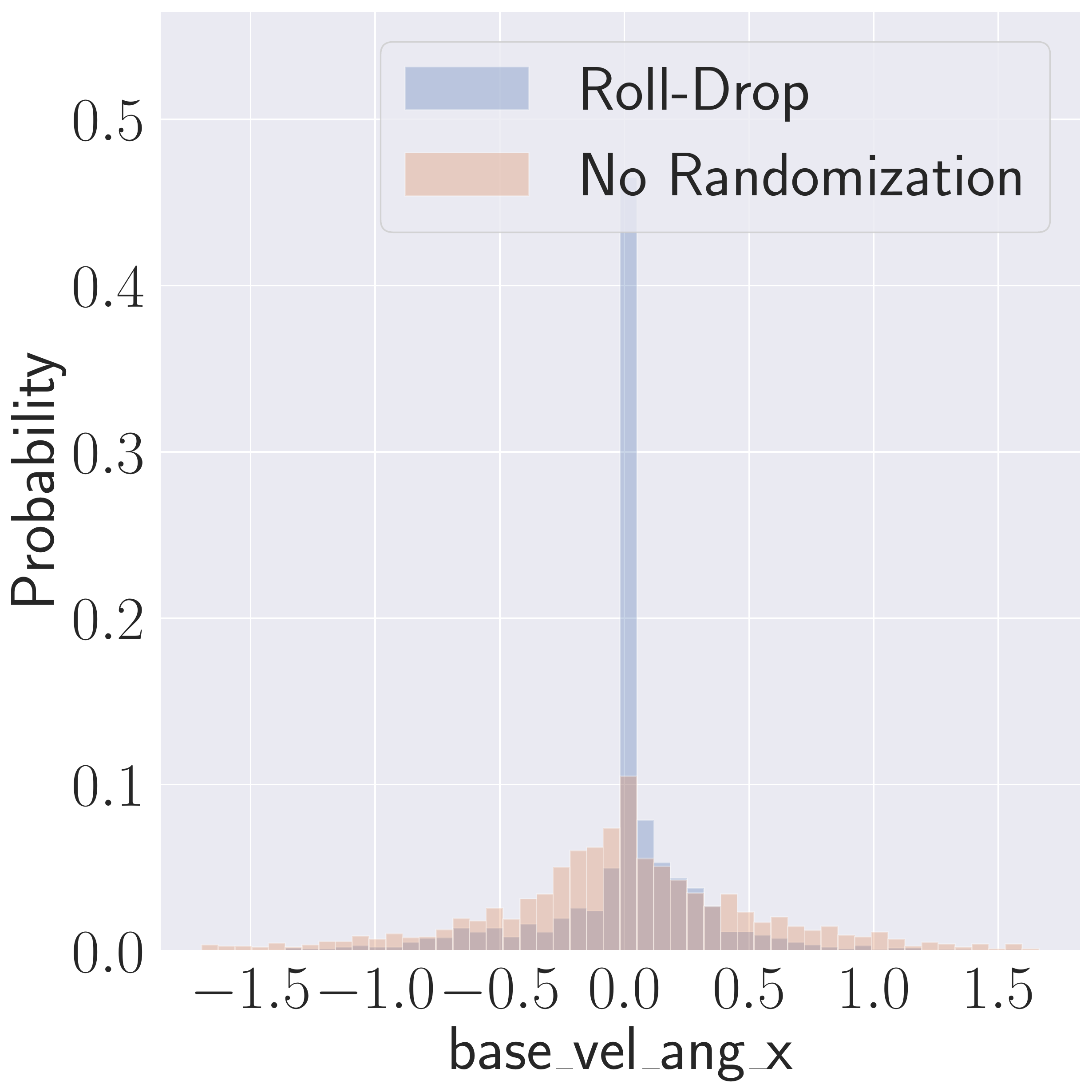}
      \label{fig:vel_ang_x}
    }
    
    \subfigure[]{
      \centering
      \includegraphics[height=3.5cm,width=0.31\textwidth]{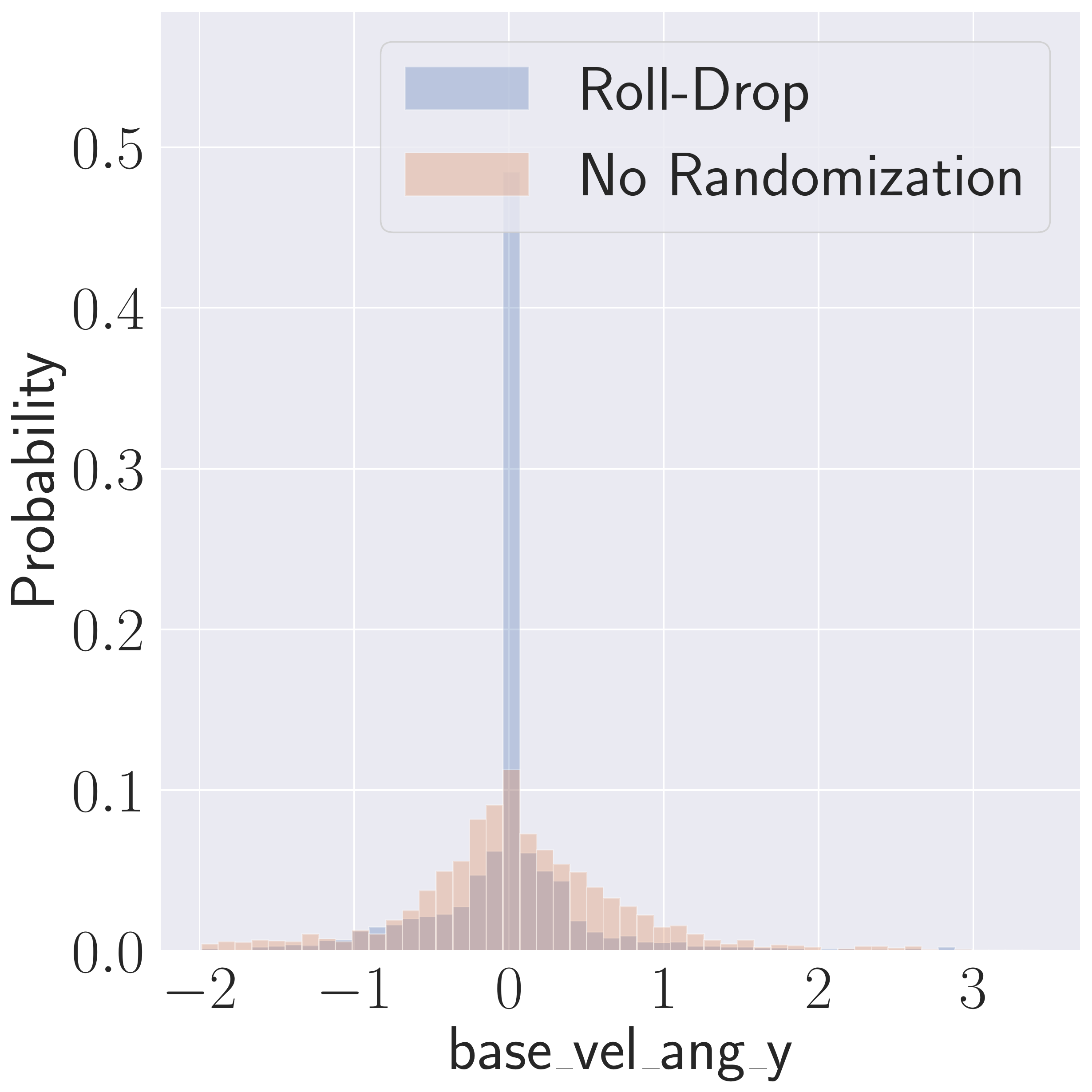}
      \label{fig:vel_ang_y}
    }
    \subfigure[]{
      \centering
      \includegraphics[height=3.5cm,width=0.31\textwidth]{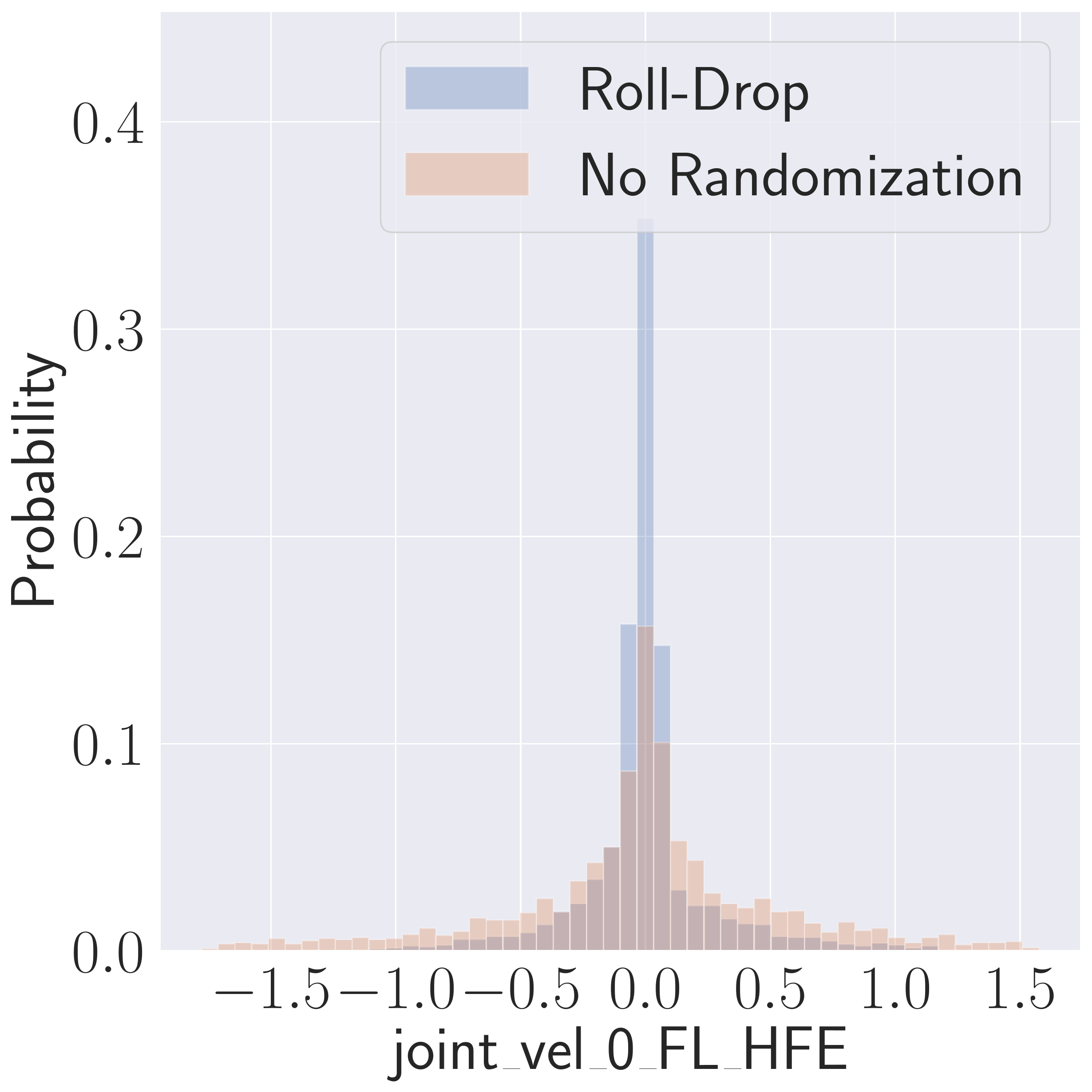}
      \label{fig:joint_vel_FL_HFE}
    }
    \subfigure[]{
      \centering
      \includegraphics[height=3.5cm,width=0.31\textwidth]{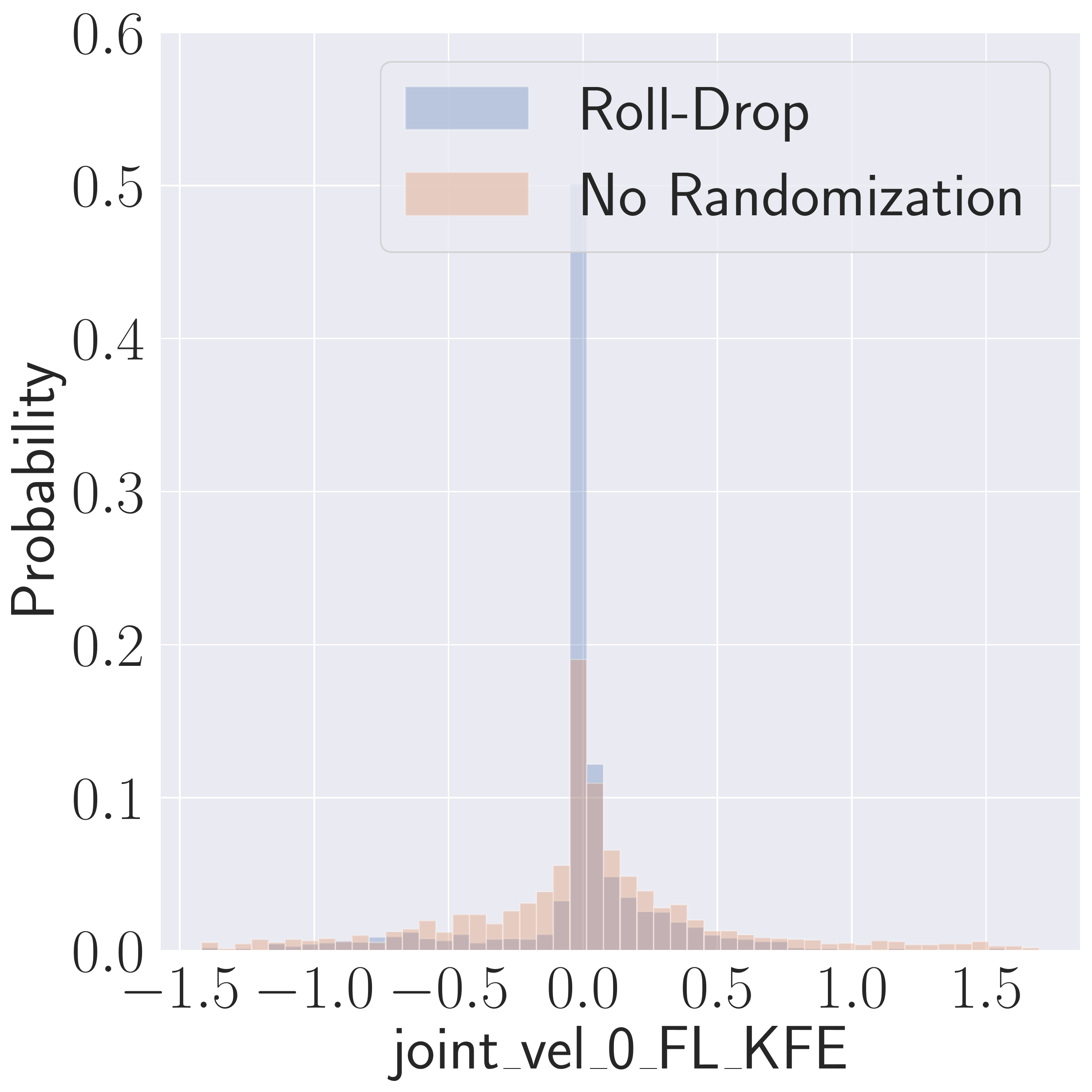}
      \label{fig:joint_vel_FL_KFE}
    }

    \caption{
    These policies are trained in simulation with $K_p=20$ and deployed on the hardware (Unitree A1) using $K_p=15$.
    In this context Roll-Drop demonstrated better performances in tracking the velocity command and in expending less velocity at the joints.
    }
    \label{fig:hardware_experiment_quantitative} 
\end{figure}

Apart from increasing the robustness to observation noise, Roll-Drop is also providing the policies with additional flexibility to external perturbations.
Indeed, we trained two more policies -with and without Roll-Drop- with $K_p=20$ and we deployed them on the hardware using $K_p=15$, the target velocity command is $s_c = [0., 0., 0.]$.
As can be seen from \Cref{fig:hardware_experiment}, the policy trained without any sort of randomisation is not able to stand, while on the other hand Roll-Drop allows the policy to find equilibrium and to better follow the velocity command.
Quantitative advantages of adopting Roll-Drop for the experiments above are provided in \Cref{fig:hardware_experiment_quantitative}, where we show better velocity command tracking (both linear and angular), and lower joint velocity usage.

\section{Conclusion} \label{sec:conclusion}
In this work we show how to account for observation noise without tuning randomisation distributions for each of the states/sensors as is commonly used in DRL.
This can be simply done by including dropout during rollouts (Roll-Drop) in the network architecture and by tuning a single parameter: the dropout probability.
In fact, by turning on and off neurons during the rollouts we show a considerable improvements in noise-injection robustness (200\%), and a success rate of 80\% when 25\% noise in injected.
Alongside the results we present a thorough analysis to explain the effects of different dropout implementations and associated probabilities on performances, convergence, and state/action distributions.
The approach was also validated on the hardware and tested on board of the Unitree A1 quadruped robot.



\clearpage

\bibliography{bibliography}

\end{document}